\documentclass[runningheads]{llncs}

\usepackage[T1]{fontenc}
\usepackage{graphicx}
\usepackage{booktabs}
\usepackage{amsmath}
\usepackage{url}
 
\newcommand{\spk}{\texttt{speaker\_proxy}}

\begin{document}

\title{Understanding Personalization Gains in Whisper ASR}
\titlerunning{Personalization Gains in Whisper ASR}

\author{Sergei Katkov\orcidID{0009-0006-1000-295X} \and Antonio Liotta\orcidID{0000-0002-2773-4421}}
\authorrunning{S. Katkov and A. Liotta}
\institute{
Free University of Bozen-Bolzano, Bolzano, Italy\\
\email{\{Sergei.Katkov, Antonio.Liotta\}@unibz.it}
}

\maketitle

\begin{abstract}
Whisper-style ASR models provide strong zero-shot transcription, but their residual errors are not evenly distributed across datasets, recording conditions, and metadata-based groups. We study when lightweight low-rank adaptation (LoRA) improves Whisper and whether observed gains require group-specific adapters.
Across Common Voice, LibriSpeech, and VoxPopuli, we compare zero-shot Whisper, pooled LoRA, and group-specific LoRA under matched within-group splits. We find that adaptation is not uniformly beneficial: gains are concentrated in difficult Common Voice groups, are consistent on VoxPopuli, and largely disappear on low-error LibriSpeech. Crucially, pooled LoRA often matches, and sometimes clearly outperforms, group-specific LoRA. Gains over zero-shot should therefore be tested against matched pooled adaptation before being interpreted as evidence of personalization.

\end{abstract}


\section{Introduction}

Whisper-style encoder-decoder ASR models, exemplified by Whisper \cite{radford2023whisper}, have made robust zero-shot transcription practical across many languages and domains. However, their residual errors are not uniformly distributed.
Recent benchmarking work demonstrated that Whisper's robustness varies substantially across noise types, signal distortions, and languages, despite strong average performance under clean conditions \cite{katkov2026robustness}.
Particular users, microphones, accents, recording environments, or speaking styles may still experience systematic failures. Personalization aims to reduce this residual mismatch by adapting the recognizer using a small amount of data from the target user, speaker, or group \cite{bell2021adaptation,meng2019speaker}.

For modern pretrained ASR, the key question is no longer whether fine-tuning can improve an under-trained recognizer. Instead, the question is whether a compact adapter, such as a low-rank adaptation (LoRA) module \cite{hu2022lora,song2024lorawhisper}, provides reliable marginal value on top of a strong general-purpose model. This distinction matters because personalization has practical costs: collecting adaptation audio, storing additional parameters, selecting the correct adapter at inference time, and avoiding degradation for groups where zero-shot recognition is already accurate. 

Another issue is that the average word error rate (WER) may obfuscate these trade-offs, because substantial gains for difficult groups may be mixed with negligible changes or small losses for easier groups.
We therefore treat personalization as a conditional intervention. Its value depends on whether the zero-shot model still makes enough systematic errors within the target group for adaptation data to be informative. If a dataset is clean, homogeneous, or already low-WER, adaptation may provide little benefit, or may even overfit. \textit{Per contra}, when the errors are concentrated in consistent group- or subset-specific conditions, a lightweight LoRA adapter may be useful \cite{joseph2024speaker,zhao2024saml}.

A further complication is that public ASR datasets do not always provide verified speaker identity. Common Voice includes an anonymized \texttt{client\_id}, which may not map cleanly to one real speaker \cite{ardila2020commonvoice}. We therefore evaluate a metadata-group personalization setting: examples from a dataset-provided group are used to adapt recognition on held-out examples from the same group, without claiming that the group is a verified identity.

Our study asks two linked questions. First, under what dataset and group conditions does LoRA adaptation improve Whisper? Second, when improvement appears, is it truly group-specific, or can a pooled adapter trained on the same selected data capture the same shared mismatch? This pooled comparison is central to our evaluation. Without it, any gain over zero-shot risks being misread as personalization, even if the adapter is only learning shared dataset, transcript, or recording conditions.

We evaluate these questions on three English speech datasets: Common Voice \cite{ardila2020commonvoice}, LibriSpeech \cite{panayotov2015librispeech}, and VoxPopuli \cite{wang2021voxpopuli}. Common Voice is crowd-sourced and heterogeneous in speaker and recording conditions; LibriSpeech is clean read audiobook speech; VoxPopuli contains parliamentary speech with public-speaking style and speaker metadata. For each selected group, we split examples into adaptation, development, and held-out test portions. All reported results should therefore be interpreted as personalization-protocol results rather than official benchmark scores.

We make the following contributions:
\begin{enumerate}
    \item We provide matched comparisons of zero-shot Whisper, pooled LoRA, and group-specific LoRA. This allows us to separate gains that require group-specific adapters from gains due to shared dataset or subset adaptation.

    \item We identify when adaptation is beneficial by combining per-group zero-shot WER analysis, repeated matched subsets, and Common Voice percentile difficulty bands. This allows us to separate cases where LoRA improves recognition from cases where pooled adaptation is sufficient.

    \item We show that improvements over zero-shot should not by themselves be interpreted as evidence of individual personalization, because pooled LoRA often matches or outperforms group-specific LoRA.
\end{enumerate}

\section{Related Work}

Speaker adaptation is a long-standing ASR problem. Classical Hidden Markov Model based (HMM) systems used transformations such as maximum likelihood linear regression (MLLR) to adapt acoustic-model parameters from small amounts of speaker data
\cite{leggetter1995mllr}. Neural acoustic models later introduced lighter speaker- or environment-dependent parameterizations, including learning hidden unit contributions (LHUC), which rescales hidden units with a small number of adapted parameters \cite{swietojanski2016lhuc}. Surveys of neural ASR adaptation organize this literature around speaker, domain, and accent adaptation, with methods based on embeddings, parameter updates, and data augmentation \cite{bell2021adaptation}. These earlier methods establish the basic motivation for adaptation: users and acoustic conditions can differ in ways that a single global recognizer does not fully absorb.

End-to-end and pretrained recognizers change the adaptation question because acoustic modeling, lexical modeling, and sequence prediction are optimized jointly. Prior work on attention-based encoder-decoder ASR studied supervised and unsupervised speaker adaptation with regularization to keep target-speaker updates near the speaker-independent model \cite{meng2019speaker}. Other end-to-end work showed that adapting small parts of a conformer-transducer model, such as attention and bias parameters, can provide rapid speaker adaptation \cite{huang2021rapid}. Large pretrained systems such as Whisper \cite{radford2023whisper} raise the bar further: for many groups the zero-shot model is already strong, so the relevant question is the marginal value of adaptation over a robust pretrained baseline.

This shift makes parameter-efficient adaptation especially important. Storing a full model per speaker or group is usually impractical, so recent ASR work has explored compact updates. Residual adapters have been used for atypical and accented speech while updating less than one percent of model parameters \cite{tomanek2021residual}. LoRA provides a general low-rank parameterization of weight updates \cite{hu2022lora}, and Whisper-focused work has used LoRA for extensible multilingual ASR adaptation \cite{song2024lorawhisper}. Closely related speaker-personalization studies use low-rank or weight-decomposed low-rank updates for end-to-end ASR \cite{joseph2024speaker}, speaker adaptation for quantized end-to-end models \cite{zhao2024quantised}, and speaker-adaptive mixtures of LoRA experts \cite{zhao2024saml}. While these studies show that compact adapters can be effective, our study asks a different control question: whether group-specific adapters outperform a pooled adapter trained on the same selected data.

Recent personalization work also shows that adaptation is most valuable under mismatch or limited coverage. Synthetic-data approaches use controllable speech synthesis and find personalization useful when the target speaker is underrepresented in global training data or when model capacity is limited \cite{yang2023text}. Other efficient speaker-adaptation methods use gated speaker-memory adapters to preserve general ASR performance while adapting to a target speaker \cite{gu2024personality}. Studies of atypical, impaired, or non-normative speech address settings where the mismatch with general-population ASR can be severe, including hypernetwork-generated updates \cite{muller2024hypernetworks}, variational LoRA for impaired speech \cite{pokel2025variational}, and two-stage initialization before speaker-specific fine-tuning \cite{jiang2026twostage}. Complementary systems address online, incremental, or multi-talker scenarios, including PI-Whisper for adaptive edge ASR \cite{nassereldine2024piwhisper} and self-speaker adaptation for multi-talker ASR \cite{wang2025speaker}. Together, these works motivate personalization, but they do not by themselves determine when a strong zero-shot model still makes enough systematic errors for group-specific LoRA to be worthwhile.

Dataset structure is therefore central to our evaluation. Common Voice \cite{ardila2020commonvoice}, LibriSpeech \cite{panayotov2015librispeech}, and VoxPopuli \cite{wang2021voxpopuli} differ in recording conditions, speaker distribution, speaking style, and transcript conventions. We use these differences to study whether personalization opportunity is better predicted by per-group WER dispersion and difficult-group tails than by average WER alone. Our contribution is complementary to mechanism-focused adaptation work: we keep LoRA simple, compare zero-shot, pooled, and group-specific adaptation under matched subsets, and use the pooled baseline to avoid over-attributing shared dataset or subset mismatch to individualized group information.

\section{Experimental Design}

\subsection{Metadata Groups for Personalization}

We use \spk{} as a compact name for the dataset-provided grouping variable:

\begin{itemize}
    \item Common Voice 24.0: \texttt{client\_id} is renamed to \spk{}.
    \item LibriSpeech: reader ID is renamed to \spk{}.
    \item VoxPopuli: \texttt{speaker\_id} is renamed to \spk{}.
\end{itemize}

This notation avoids treating Common Voice \texttt{client\_id} as ground-truth speaker identity. It also narrows the experimental claim: we evaluate whether metadata-consistent groups support adaptation, not whether the metadata perfectly identify real speakers.

\subsection{Evaluation Splits}

For all datasets, examples are split into train/dev/test portions inside each selected group after excluding entries without transcript, audio, or group metadata. The default ratio is 60/20/20. This differs from using official dataset splits directly, but it matches the personalization setting: a small amount of adaptation data from a group is used to evaluate recognition on held-out utterances from the same group.

The split is performed independently within each group, so every selected group contributes both adaptation examples and held-out evaluation examples. This avoids a mismatch with official train/test partitions, which often contain disjoint speakers and therefore test speaker-independent generalization rather than same-group adaptation. Our scores are personalization-protocol results, not official benchmark scores.

For adaptation, we report matched repeated subsets. For a given dataset, model, and random seed, zero-shot, pooled LoRA, and personalized LoRA are evaluated on the same groups and the same held-out utterances. This makes the comparison local to the adapted groups: gains are computed against the zero-shot WER of the same selected groups, not against a separate full-population average.

\subsection{Compared Systems}

We compare three methods:

\begin{itemize}
    \item \textbf{Zero-shot Whisper}: pretrained Whisper evaluated without adaptation.
    \item \textbf{Pooled LoRA}: one LoRA adapter trained on data pooled across selected groups.
    \item \textbf{Personalized LoRA}: one LoRA adapter trained separately for each group.
\end{itemize}

For Common Voice, we also include an auxiliary pooled LoRA check over a broader group population. It probes broad Common Voice mismatch, but it is not a personalization comparison because it has no matched personalized control.

The adaptation budget is 200 training utterances for each Common Voice group and 80 training utterances for each LibriSpeech or VoxPopuli group. Because Common Voice utterances are shorter, this gives a comparable amount of adaptation audio across datasets. Mean utterance duration is 5.1 s for Common Voice, 12.3 s for LibriSpeech, and 10.0 s for VoxPopuli.

For personalized LoRA, the budget is applied separately to each group adapter. For pooled LoRA, the corresponding training examples from all selected groups are concatenated into one adaptation set. The pooled baseline controls for exposure to the same dataset, transcript style, and recording conditions while removing the per-group adapter choice at inference time.

\subsection{Models and Training Details}

We evaluate the OpenAI Whisper tiny, base, small, and medium.
We adapt Whisper with LoRA updates on the attention projections, following the parameter-efficient adaptation rationale of LoRA and recent Whisper adaptation work \cite{hu2022lora,song2024lorawhisper}. LoRA keeps the pretrained weights frozen and learns a low-rank update
\[
    W' = W + \Delta W,\qquad \Delta W = BA,
\]
where $A$ and $B$ have rank much smaller than the original weight matrix. In our runs, LoRA is applied to the query and value projections with rank 16, scaling factor 32, and dropout 0.1. Training is monitored with development-set WER and uses early stopping; the checkpoint with the best development WER is evaluated. Each reported adaptation result is averaged over five independent runs with different sampled group subsets. Within each run, zero-shot, pooled LoRA, and personalized LoRA are evaluated on the same held-out utterances.

Before scoring, references and hypotheses are normalized consistently within each dataset by lowercasing text and removing punctuation and other non-lexical formatting differences. Recognition quality is measured with word error rate (WER),
\[
    \mathrm{WER} = \frac{S + D + I}{N},
\]
where $S$, $D$, and $I$ are the numbers of substituted, deleted, and inserted words after alignment, and $N$ is the number of reference words. WER values are reported as percentages. The main adaptation metric is absolute WER gain, defined as zero-shot WER minus adapted WER on the matched test subset; positive values indicate improvement. Repeated-subset results are reported as mean $\pm$ 95\% confidence intervals over the five sampled group subsets. For zero-shot columns in matched adaptation tables, the interval reflects subset-selection variability rather than stochastic decoding.

\subsection{Common Voice Difficulty Analysis}

For Common Voice, we additionally evaluate percentile-stratified variants of the adaptation experiment. For each Whisper model size, eligible groups are ranked by their zero-shot WER on the held-out split. We then form model-specific difficulty subsets: the hardest 10\%, hardest 20\%, middle 40--60\%, and easiest 20\%. This design avoids comparing fixed WER thresholds across models with very different baseline accuracies. It also tests whether adaptation gains shrink as the zero-shot baseline becomes more accurate within the same dataset and model family.

We use this ranking in two ways. First, a pooled LoRA sweep shows how adaptation changes across the WER distribution. Second, matched repeated analyses compare pooled and group-specific adapters within the same difficulty bands. This tests whether personalization adds value beyond adapting to a difficult subset.

\section{Datasets}

Table~\ref{tab:dataset-stats} summarizes the data available under our within-group protocol. These are not official benchmark splits; adaptation experiments sample smaller repeated subsets from these populations.

\begin{table}
\caption{Dataset statistics under the within-group protocol. Audio hours are approximate totals; repeated adaptation subsets are sampled separately.}
\label{tab:dataset-stats}
\centering
\renewcommand{\arraystretch}{1.08}
\begin{tabular*}{\textwidth}{@{\extracolsep{\fill}}lrrrr@{}}
\toprule
Dataset & \shortstack{Groups} & Utterances & \shortstack{Approx. audio\\hours} & \shortstack{Mean duration\\(s)} \\
\midrule
Common Voice & 9,139 & 1,376,296 & 1,932.7 & 5.06 \\
LibriSpeech & 1,949 & 249,787 & 852.9 & 12.29 \\
VoxPopuli & 260 & 106,779 & 295.8 & 9.97 \\
\bottomrule
\end{tabular*}
\end{table}

\paragraph{Common Voice.}
Common Voice English is crowd-sourced speech with diverse speakers, recording devices, and environments. We use the anonymized \texttt{client\_id} as the grouping variable and create within-group splits.

\paragraph{LibriSpeech.}
LibriSpeech is clean read English audiobook speech with reader identifiers. For same-group personalization, we create within-reader splits rather than using the official speaker-disjoint partitions.

\paragraph{VoxPopuli.}
VoxPopuli English ASR contains parliamentary speech with public-speaking style and speaker metadata. We use \texttt{speaker\_id} as the grouping variable and create within-group splits.

\section{Per-Group WER Distributions}

Our hypothesis is that personalization helps when baseline errors are sufficiently large and group-structured.

Table~\ref{tab:difficulty} reports per-group zero-shot WER distributions for Whisper small. The mean, median, P90 (90th percentile), and P99 (99th percentile) columns summarize the distribution of group-level WER values. The last three columns report tail mass: the percentage of groups whose zero-shot WER is above a fixed threshold. 
For example, the Common Voice value of 25.5 in the WER $>$10\% column means that 25.5\% of Common Voice groups have zero-shot WER higher than 10\%.

\begin{table}
\caption{Per-group zero-shot WER distributions with Whisper small. WER values and tail masses are expressed as percentages; P90 and P99 denote the 90th and 99th percentiles across groups.}
\label{tab:difficulty}
\centering
\renewcommand{\arraystretch}{1.08}
\begin{tabular*}{\textwidth}{@{\extracolsep{\fill}}lrrrrrrr@{}}
\toprule
Dataset & \shortstack{Mean\\WER} & \shortstack{Median\\WER} & \shortstack{P90\\WER} & \shortstack{P99\\WER} & \shortstack{WER\\$>$10\%} & \shortstack{WER\\$>$15\%} & \shortstack{WER\\$>$20\%} \\
\midrule
Common Voice & 7.89 & 6.42 & 15.28 & 31.58 & 25.5 & 10.4 & 4.4 \\
LibriSpeech & 5.01 & 4.45 & 8.43 & 13.85 & 4.8 & 0.6 & 0.3 \\
VoxPopuli & 11.37 & 10.82 & 15.23 & 22.75 & 63.8 & 10.4 & 2.3 \\
\bottomrule
\end{tabular*}
\end{table}

\begin{figure}
\centering
\includegraphics[width=0.88\textwidth]{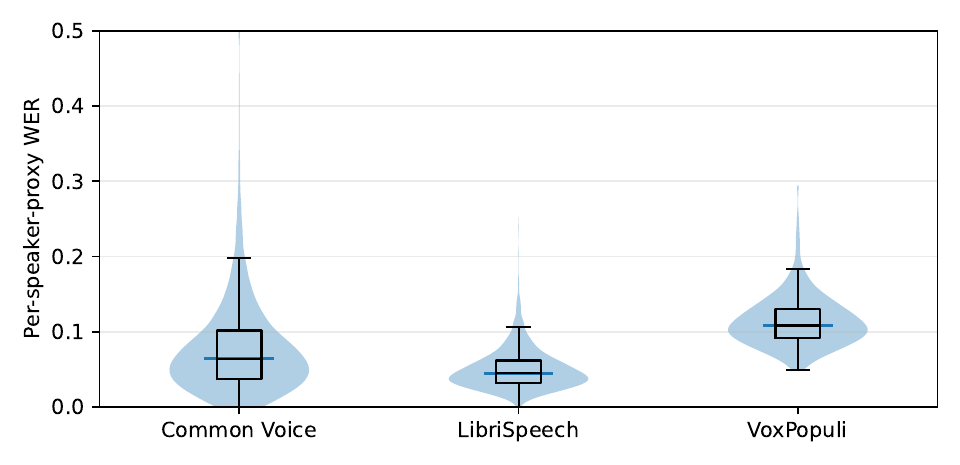}
\caption{Per-group zero-shot WER distributions for Whisper small. The y-axis shows WER as a fraction. The violin shapes summarize the distribution across speaker proxies, and the box plots indicate the central tendency and spread. The long upper tails highlight difficult-group effects, especially for Common Voice.}
\label{fig:wer-distributions}
\end{figure}

The datasets differ not only in average WER, but also in how that error is distributed. LibriSpeech is easy and comparatively homogeneous. Common Voice has a pronounced tail of difficult groups. VoxPopuli is harder on average and has many moderately difficult groups, but its tail is less extreme than Common Voice. Figure~\ref{fig:wer-distributions} visualizes this contrast.

\section{Results}

\subsection{Overall Zero-Shot Landscape}

Table~\ref{tab:zeroshot} shows zero-shot WER percentages on the full evaluation population defined by the within-group protocol: 9,139 groups for Common Voice, 1,949 for LibriSpeech, and 260 for VoxPopuli. These numbers give broad dataset context; they are not official benchmark scores or the matched subsets used for adaptation.

\begin{table}
\caption{Overall zero-shot Whisper performance under the within-group protocol. WER values are percentages; these population-level numbers provide context for the matched adaptation subsets in Table~\ref{tab:cross-dataset-adaptation}.}
\label{tab:zeroshot}
\centering
\renewcommand{\arraystretch}{1.08}
\begin{tabular*}{\textwidth}{@{\extracolsep{\fill}}llr@{}}
\toprule
Dataset & Model & WER\% \\
\midrule
Common Voice & tiny & 18.71 \\
Common Voice & base & 13.76 \\
Common Voice & small & 9.02 \\
Common Voice & medium & 6.89 \\
\midrule
LibriSpeech & tiny & 9.65 \\
LibriSpeech & base & 6.92 \\
LibriSpeech & small & 5.01 \\
LibriSpeech & medium & 5.36 \\
\midrule
VoxPopuli & tiny & 15.25 \\
VoxPopuli & base & 12.83 \\
VoxPopuli & small & 11.10 \\
VoxPopuli & medium & 9.97 \\
\bottomrule
\end{tabular*}
\end{table}

The overall zero-shot results scale cleanly with model size on Common Voice and VoxPopuli. LibriSpeech is already low-WER for all evaluated models and leaves little room for adaptation; in this protocol, the small model is slightly better than medium on LibriSpeech. We use LibriSpeech mainly as a low-error control rather than as a scaling study. The adaptation comparisons below use matched repeated subsets, so zero-shot, pooled LoRA, and personalized LoRA are compared on the same selected groups and held-out utterances.

\subsection{Adaptation Across Datasets}

We first compare zero-shot Whisper, pooled LoRA, and personalized LoRA across the three datasets. Table~\ref{tab:cross-dataset-adaptation} reports matched repeated-subset zero-shot WER alongside pooled and personalized LoRA WER.

\begin{table}
\caption{Adaptation across datasets. WER values are percentages; bold marks the best adapted condition in each row.}
\label{tab:cross-dataset-adaptation}
\centering
\small
\setlength{\tabcolsep}{5pt}
\renewcommand{\arraystretch}{1.12}
\begin{tabular*}{\textwidth}{@{\extracolsep{\fill}}llrrr@{}}
\toprule
Dataset & Model & \shortstack{Matched\\zero-shot WER} & \shortstack{Pooled\\LoRA WER} & \shortstack{Personalized\\LoRA WER} \\
\midrule
Common Voice & tiny & 19.66 $\pm$ 2.85 & 19.31 $\pm$ 0.95 & \textbf{18.69 $\pm$ 1.12} \\
Common Voice & base & 14.29 $\pm$ 2.11 & 14.13 $\pm$ 0.77 & \textbf{14.10 $\pm$ 0.62} \\
Common Voice & small & 9.38 $\pm$ 1.64 & \textbf{9.12 $\pm$ 0.53} & 9.47 $\pm$ 0.50 \\
Common Voice & medium & 7.11 $\pm$ 1.40 & \textbf{6.75 $\pm$ 0.50} & 6.94 $\pm$ 0.55 \\
\midrule
LibriSpeech & tiny & 11.62 $\pm$ 1.20 & \textbf{11.61 $\pm$ 1.09} & 11.65 $\pm$ 1.24 \\
LibriSpeech & base & 8.14 $\pm$ 0.58 & \textbf{8.05 $\pm$ 0.70} & 8.09 $\pm$ 0.68 \\
LibriSpeech & small & 5.54 $\pm$ 0.30 & 5.59 $\pm$ 0.39 & \textbf{5.51 $\pm$ 0.35} \\
LibriSpeech & medium & 6.36 $\pm$ 0.54 & 6.74 $\pm$ 0.88 & \textbf{6.17 $\pm$ 0.58} \\
\midrule
VoxPopuli & tiny & 15.30 $\pm$ 0.35 & \textbf{14.18 $\pm$ 0.33} & 14.20 $\pm$ 0.24 \\
VoxPopuli & base & 12.93 $\pm$ 0.70 & \textbf{11.27 $\pm$ 0.29} & 12.02 $\pm$ 0.19 \\
VoxPopuli & small & 11.13 $\pm$ 0.38 & \textbf{8.65 $\pm$ 0.29} & 10.37 $\pm$ 0.26 \\
VoxPopuli & medium & 10.24 $\pm$ 0.43 & 9.81 $\pm$ 0.41 & \textbf{9.80 $\pm$ 0.43} \\
\bottomrule
\end{tabular*}
\end{table}

The cross-dataset results do not reduce to a single win/loss pattern. VoxPopuli benefits most consistently, especially from pooled LoRA for the base and small models. LibriSpeech behaves like a low-error control: both pooled and personalized adapters remain close to the matched zero-shot baseline, with negligible gains or losses. Common Voice is harder to summarize with one average because the matched subsets combine easy and difficult groups, so we next separate it by baseline difficulty.

The matched zero-shot column should be read separately from the overall zero-shot table. Adaptation requires enough same-group audio, so repeated subsets are drawn from eligible groups rather than from the entire population. Common Voice and LibriSpeech are also stratified by zero-shot WER to avoid sampling only easy or difficult groups. This distinction is strongest for LibriSpeech, where the eligibility filter favors harder readers. Thus Table~\ref{tab:zeroshot} describes the broader population, while Table~\ref{tab:cross-dataset-adaptation} gives the matched baseline for the adaptation comparison.

\subsection{Adaptation Gains on Common Voice}

Table~\ref{tab:cv-percentiles} shows a pooled LoRA sweep over Common Voice difficulty percentiles. Within each model size, proxies are ranked by zero-shot WER and grouped into hard, middle, and easy bands. The absolute WER reduction is largest in the high-error tail and becomes small for easier groups. Percentage-point reduction is the most direct measure of adaptation benefit in these high-error groups.

\begin{table}
\caption{Pooled LoRA on Common Voice difficulty bands. WER columns are percentages; gain is the absolute percentage-point improvement over zero-shot.}
\label{tab:cv-percentiles}
\centering
\small
\setlength{\tabcolsep}{5pt}
\renewcommand{\arraystretch}{1.12}
\begin{tabular*}{\textwidth}{@{\extracolsep{\fill}}llrrr@{}}
\toprule
Model & \shortstack{Difficulty\\band} & \shortstack{Zero-shot\\WER} & \shortstack{Pooled\\WER} & \shortstack{Gain\\(pp)} \\
\midrule
tiny & Hardest 10\% & 48.81 & 38.94 & +9.88 \\
base & Hardest 10\% & 38.00 & 31.30 & +6.70 \\
small & Hardest 10\% & 26.29 & 22.21 & +4.09 \\
medium & Hardest 10\% & 19.71 & 17.49 & +2.22 \\
\midrule
tiny & Hardest 20\% & 38.13 & 32.44 & +5.69 \\
base & Hardest 20\% & 29.02 & 24.93 & +4.09 \\
small & Hardest 20\% & 19.31 & 17.47 & +1.83 \\
medium & Hardest 20\% & 14.43 & 13.48 & +0.95 \\
\midrule
tiny & Middle 40--60\% & 14.86 & 13.00 & +1.86 \\
base & Middle 40--60\% & 10.59 & 9.59 & +1.00 \\
small & Middle 40--60\% & 6.86 & 6.20 & +0.66 \\
medium & Middle 40--60\% & 5.32 & 4.73 & +0.58 \\
\midrule
tiny & Easiest 20\% & 7.77 & 7.55 & +0.22 \\
base & Easiest 20\% & 5.35 & 4.87 & +0.48 \\
small & Easiest 20\% & 3.41 & 3.01 & +0.40 \\
medium & Easiest 20\% & 2.62 & 2.27 & +0.35 \\
\bottomrule
\end{tabular*}
\end{table}

The hardest-10\% rows make the dependence on baseline difficulty visible. For Whisper medium, the full Common Voice zero-shot WER is 6.89\%, yet the hardest eligible groups have a zero-shot WER of 19.71\% and gain 2.22 percentage points from pooled LoRA. The medium hardest-20\% band gains only 0.95 points. At the easy end, the absolute gains are much smaller: tiny improves by 0.22 points in the easiest-20\% band versus 9.88 points in the hardest-10\% band, and base improves by 0.48 points versus 6.70 points. Small and medium still improve in the easiest-20\% rows, but by less than half a point. Adaptation value is tied to local baseline difficulty, not only to dataset-average WER.

\subsection{Pooled and Personalized Adaptation in Difficult Groups}

The pooled sweep shows where adaptation is beneficial, but not whether the adapter needs to be group-specific. Table~\ref{tab:cv-difficulty-repeats} compares pooled and personalized LoRA within the same Common Voice difficulty bands. To keep the table focused, we report gains rather than adapted WER.

The two Common Voice analyses answer different questions. 
Table~\ref{tab:cv-percentiles} measures how pooled adaptation varies across fixed zero-shot difficulty bands, whereas Table~\ref{tab:cv-difficulty-repeats} uses matched repeated subsets to compare pooled and group-specific adapters on the same sampled groups and held-out utterances. 
Thus, Table~\ref{tab:cv-difficulty-repeats} is not intended to reproduce the pooled sweep in Table~\ref{tab:cv-percentiles} exactly. 
Instead, it tests whether group-specific adapters provide additional benefit beyond a matched pooled adapter within the same difficulty regime.

\begin{table}
\caption{Common Voice difficulty-band comparison of pooled and personalized LoRA. Values are absolute WER gains in percentage points.}
\label{tab:cv-difficulty-repeats}
\centering
\small
\setlength{\tabcolsep}{5pt}
\renewcommand{\arraystretch}{1.12}
\begin{tabular*}{\textwidth}{@{\extracolsep{\fill}}llrrr@{}}
\toprule
Model & Method & \shortstack{Hardest\\10\%} & \shortstack{Middle\\40--60\%} & \shortstack{Easiest\\20\%} \\
\midrule
tiny & Pooled & \textbf{+12.61 $\pm$ 0.44} & +1.47 $\pm$ 0.11 & +0.35 $\pm$ 0.06 \\
tiny & Personalized & +11.36 $\pm$ 0.22 & +1.43 $\pm$ 0.05 & \textbf{+0.37 $\pm$ 0.04} \\
\midrule
base & Pooled & \textbf{+10.91 $\pm$ 0.19} & \textbf{+0.87 $\pm$ 0.02} & \textbf{+0.43 $\pm$ 0.02} \\
base & Personalized & +8.77 $\pm$ 0.15 & +0.59 $\pm$ 0.07 & +0.15 $\pm$ 0.03 \\
\midrule
small & Pooled & \textbf{+8.08 $\pm$ 0.24} & \textbf{+0.62 $\pm$ 0.05} & \textbf{+0.51 $\pm$ 0.03} \\
small & Personalized & +6.22 $\pm$ 0.24 & +0.17 $\pm$ 0.03 & +0.08 $\pm$ 0.02 \\
\midrule
medium & Pooled & \textbf{+6.66 $\pm$ 0.11} & \textbf{+0.70 $\pm$ 0.04} & \textbf{+0.51 $\pm$ 0.02} \\
medium & Personalized & +5.06 $\pm$ 0.12 & +0.31 $\pm$ 0.03 & +0.06 $\pm$ 0.04 \\
\bottomrule
\end{tabular*}
\end{table}

The matched difficulty runs lead to the same conclusion. The largest absolute gains occur in the hardest band: pooled LoRA improves hardest-10\% WER by 12.61, 10.91, 8.08, and 6.66 percentage points for tiny, base, small, and medium, respectively. In the middle and easiest bands, the gains are much smaller. Pooled LoRA is also stronger than personalized LoRA for all hardest-10\% rows and for most middle and easy rows. The gain should not be read as purely individual personalization; much of it is captured by adapting once to a difficult subset.

\section{Conclusion}

Our results show that Whisper adaptation should be treated as a selective intervention rather than a uniform personalization step. Using a matched within-group protocol on Common Voice, LibriSpeech, and VoxPopuli, we find that LoRA adaptation is most useful when zero-shot Whisper still makes stable, group-structured errors. VoxPopuli shows consistent benefits from pooled LoRA, while Common Voice gains are concentrated in the high-error tail and shrink for easier groups and stronger model sizes. LibriSpeech serves as a low-error control condition, with adapted models remaining close to the zero-shot baseline.

The pooled baseline is essential for interpreting these gains. Because pooled LoRA often matches or outperforms group-specific LoRA, improvements over zero-shot are not sufficient evidence of individual personalization. In several cases, the gains are better explained as adaptation to shared dataset, transcript, recording, or difficult-subset conditions. Practical personalization studies should therefore compare against matched pooled adapters, inspect per-group WER distributions, and focus adaptation on groups with measurable and stable baseline errors.

Several limitations remain. First, the grouping variable is a metadata proxy rather than verified identity, especially for Common Voice. Second, the within-group protocol produces personalization-oriented results rather than official benchmark scores. Third, the adaptation budgets are only approximately matched across datasets, and we do not separately analyze acoustic factors such as SNR, channel, or reverberation.

The Common Voice matched difficulty repeats should also be interpreted diagnostically. They cover selected percentile bands rather than all possible bands, and difficulty-band selection in deployment would need to rely on development data or historical zero-shot performance, not on held-out test performance. These limitations reinforce the main recommendation: personalization should be treated as a selective intervention, not as a uniform post-processing step.

\begin{credits}

\subsubsection{\ackname}
This work was partially supported by computational resources provided by CINECA. We acknowledge the CINECA award under the ISCRA initiative for the availability of high-performance computing resources and support.

\end{credits}

\bibliographystyle{splncs04}
\bibliography{references}

\end{document}